\documentclass{article}

\usepackage{amsmath, amssymb, amsfonts}
\usepackage{amsthm}

\newtheorem{lemma}{Lemma}
\newtheorem{theorem}{Theorem}

\theoremstyle{definition}

\newcommand{\E}{\mathbb{E}}

\usepackage{xr-hyper}
\usepackage{xr}
\usepackage{multirow}
\usepackage{microtype}
\usepackage{amsmath}
\usepackage{amsfonts}
\usepackage{graphicx}
\usepackage{colortbl}
\usepackage{caption}
\usepackage{subcaption}
\usepackage{xcolor}
\usepackage{xr-hyper}
\usepackage{booktabs} 

\DeclareMathOperator{\var}{var}
\DeclareMathOperator{\cov}{cov}
\usepackage{hyperref}

\usepackage[preprint]{neurips_2020}

\usepackage[utf8]{inputenc} 
\usepackage[T1]{fontenc}    
\usepackage{hyperref}       
\usepackage{url}            
\usepackage{booktabs}       
\usepackage{amsfonts}       
\usepackage{nicefrac}       
\usepackage{microtype}      

\title{A Semiparametric Approach to Interpretable Machine Learning}

\author{%
   Numair Sani \\
  Johns Hopkins University \\
  Baltimore, MD 21218\\
  \texttt{snumair1@jhu.edu} \\
  \And
  Jaron Lee\\
  Johns Hopkins University \\
  Baltimore, MD 21218\\
  \texttt{jaron.lee@jhu.edu} \\ 
  \AND
  Razieh Nabi \\
  Johns Hopkins University \\
  Baltimore, MD 21218\\
  \texttt{rnabi@jhu.edu} \\
  \And
  Ilya Shpitser \\
  Johns Hopkins University \\
  Baltimore, MD 21218\\
  \texttt{ilyas@cs.jhu.edu} \\
}

\date{}

\begin{document}

\maketitle

\begin{abstract}
Black-box models in machine learning have demonstrated excellent predictive performance in complex problems and high-dimensional settings. 
However, their lack of transparency and interpretability restrict the applicability of such models in critical decision-making processes. 
In order to combat this shortcoming, we propose a novel approach to trading off interpretability and performance in prediction models using ideas from semiparametric statistics, allowing us to combine the interpretability of parametric regression models with performance of nonparametric methods. We achieve this by utilizing a two-piece model: the first piece is interpretable and parametric, to which a second, uninterpretable residual piece is added. The performance of the overall model is optimized using methods from the sufficient dimension reduction literature. Influence function based estimators are derived and shown to be \textit{doubly robust}. This allows for use of approaches such as Double Machine Learning 
in estimating our model parameters. We illustrate the utility of our approach via simulation studies and a data application based on predicting the length of stay in the intensive care unit among surgery patients.
\end{abstract}

\section{Introduction}
\label{sec:intro}
Machine learning (ML) algorithms are becoming easier to train and deploy with high quality off-the-shelf software packages, and yield excellent prediction performance even in complex high-dimensional settings. A known drawback of such complex models, however, is their opaqueness; it is difficult for a human being examining the model to understand why a particular prediction was made \cite{opacityml}. In domains such as healthcare, resume screening, and recidivism prediction, decisions based on output of predictive models have critical downstream consequences or ethical ramifications.  In such cases it is crucial to understand why a particular prediction was made by the algorithm. Additionally, transparency in the underlying decision process engenders trust in these algorithms, making consumers more likely to utilize them.

Interpretable Machine Learning is a growing and active field of research. While the precise meaning of "interpretable" remains ambiguous, researchers have aimed to lay down a broad framework for interpretable ML \cite{doshivelez2017rigorous, lipton2016mythos, Murdoch_2019, du2018techniques, jamiainterpretable}. These papers outline the properties that models must possess in order to count as interpretable, put forth hierarchies and along with discussions of the merits and drawbacks of existing approaches to interpretable ML.

Broadly, there are two approaches to interpretable ML. The first approach involves developing models that possess certain interpretable properties themselves, such as rule-based models, simple parametric models and case-based models. For instance, \cite{wang2016bayesian} state that rule sets are interpretable in a certain sense, and propose a procedure for learning rule sets for the purpose of classification. \cite{choi2016retain} make use of an interpretable two-level attention model for prediction purposes. However, such models place constraints on model complexity, and often lose out on predictive performance.  

The second approach, utilized when the performance of interpretable models is inadequate involves ``post-hoc" interpretability; i.e. interpreting an already trained black-box model. Examples of such approaches saliency maps \cite{simonyan2013deep, fong2017interpretable}, producing local explanations - explanations for individual predictions \cite{ribeiro2016i} or global predictions\cite{lakkaraju2017interpretable}. However, such methods have their drawbacks as well since they only learn approximations of the underlying black-box, and it is hard to provide guarantees on the accuracy and reliability of such approximations.

A new approach growing in popularity involves combining simple interpretable models with the predictive power of uninterpretable, black-box models. For example, \cite{alvarez2018towards} combined the interpretability of coefficients in a linear regression model with the model flexibility of neural networks to yield greater performance while preserving interpretability, whilst \cite{pmlr-v97-wang19a} learned an interpretable model in place of a black-box model for subsets of the data, while leaving the rest of the black-box model intact.

In this paper, we utilize a similar approach to interpretability as above. We start with a simple parametric model relating a set of interpretable features to the outcome, and add a more complicated function consisting of uninterpretable features in a way that improves its performance as much as possible.  This function may be viewed as an ``uninterpretable residual'', augmenting a simple parametric model. We use ideas from  sufficient dimension reduction literature \cite{ma2012semiparametric} and semiparametric statistics \cite{tsiatis2007semiparametric}, to estimate such a residual under minimal assumptions about the underlying data generating process.

The paper is organized as follows.
In Section \ref{sec:tradeoff}, we put forth our reasoning on the tradeoff between interpretability and model performance and motivate the structure of our interpretable model. Section \ref{sec:sdr} describe the concepts utilized in fitting our model, namely Sufficient Dimension Reduction (SDR) and semiparametric statistics.  Section \ref{sec:inter_ML} formally describes our interpretable model, while Section \ref{sec:exp} demonstrates the performance of our model on simulated data as well as a prediction task using healthcare data. Section \ref{sec:conc} contains our conclusions.

\section{The Tradeoff Between Interpretability And Performance}
\label{sec:tradeoff}

In order to demonstrate the tradeoff between interpretability and performance, and to motivate our interpretable model, we will consider a healthcare example. We aim to develop a predictive model for Intensive Care Unit (ICU) length of stay for patients who have undergone cardiac surgery, using a detailed institutionally collected electronic health record. Accurate prediction of ICU Length of stay (LOS) is of interest to clinicians as it allows them to plan for measures such as palliative care consultation, early mobility therapy or discharge to a long-term acute care facility \cite{kramer2010predictive}. Additionally, it also allows for clinicians to target interventions described in \cite{lee2019intensive}. Prolonged ICU stays are also related to catheter related bloodstream infections\cite{pronovost2006intervention}, hence accurate prediction of LOS can allow clinicians to implement additional safeguards against healthcare associated infections (HAIs).
 
In such a setting, building a useful prediction model involves a tradeoff. One approach is to use modern ML algorithms, capable of learning complex relationships in the data and providing good empirical performance. However, such methods are often opaque, making it hard for clinicians to understand why a particular prediction was made for a particular patient and ascertain why the model prediction disagrees with current clinical consensus. Additionally, interpretable models that predict undesirable outcomes can be useful for informing practice and ultimately improving patient care.

On the other end of the spectrum, one could utilize risk scores based on relatively simple statistical models defined on a small set of clinically interpretable variables $X_{\text{int}}$ thought to strongly influence the outcome of interest.  For example, the Parsonnet score \cite{parsonnetMethodUniformStratification1989} is used to predict post cardiac surgery mortality.  It is based on a relatively small set of variables thought to influence  risk, including gender, obesity, diabetes status, the presence of hypertension, age, the number of previous operations, and so on \cite{lawrenceParsonnetScoreGood2000}, and each variable gets a certain number of points, and these points are calculated out a 100 and interpreted as a percentage. Such a risk score is an oversimplification as the set of variables truly relevant for the given outcome is much larger than what is used in risk score models.  Furthermore, it is likely that the true regression surface relating this much larger, uninterpretable set of variables $X_{\text{uint}}$ to the outcome $Y$ is very complicated.

Our approach aims to bridge the gap between the true (but likely uninterpretable) regression surface $\mathbb{E}_0[Y | X_{\text{uint}}, X_{\text{int}}]$ and the interpretable (but likely inaccurate) regression surface $\mathbb{E}_{\text{int}}[Y | X_{\text{int}}]$ used in practice.
To accomplish this, our approach starts with a simple model $\mathbb{E}_{\text{int}}[Y | X_{\text{int}}]$, such as a linear regression $X_{\text{int}}^T \psi$, and augments it with an \emph{unrestricted} function $r()$ of a low dimensional version $X^*_{\text{uint}}$ of the uninterpretable set of features $X_{\text{uint}}$, obtained via a function $g : X_{\text{uint}} \mapsto X^*_{\text{uint}}$ known up to a finite set of parameters $\gamma$.
We do so in such a way that the resulting surface $X_{\text{int}}^T  \psi + r(g(X_{\text{uint}}; \gamma))$ is close to the true surface $\mathbb{E}_0[Y | X_{\text{uint}}, X_{\text{int}}]$.

Our task is to find the parameter set $\beta \equiv (\psi, \gamma)$ that best captures $\mathbb{E}_0[Y | X_{\text{uint}}, X_{\text{int}}]$ given a set of realizations of the observed data distribution $p_0(Y, X_{\text{uint}}, X_{\text{int}})$.  
Successfully solving our task yields a surface that does a much better job of capturing features of the true surface $\mathbb{E}_0$, while retaining much of the interpretable structure of $\mathbb{E}_{\text{int}}$.  In particular, we can view $r()$ as a kind of low-dimensional uninterpretable residual that when added to an interpretable model improves performance.  This addition allows the subject matter expert to explicitly examine cases where a simpler model $\mathbb{E}_{\text{int}}$ differs from a better performing one by examining the behavior of $r()$, and we provide examples examining this in our experiments section. We emphasize again that $r()$ is completely unrestricted.

We approach the problem of learning $\beta$ using ideas from the sufficient dimension reduction (SDR) literature and semiparametric statistics.

\section{Sufficient Dimension Reduction and Semiparametic Statistics}
\label{sec:sdr}


Sufficient Dimension Reduction refers to a popular class of methods that perform dimension-reduction while taking the feature outcome relationships into account. These methods are different from those such as Principal Components Analysis (PCA)\cite{wold1987principal}, which ignore the relationships between features and outcomes. A rich literature exists in statistics on sufficient dimension reduction (SDR), discussed in the Supplement. Broadly, the SDR problem is set up as:

Let $X$ be a $p$-dimensional covariate vector and $Y$ be a univariate response variable. The goal of SDR is to find a known function $r(.; \gamma)$, with a much smaller range than the domain, parameterized by $\gamma$ such that $\E_0[Y \mid X] = \E[Y \mid r(X; \gamma)],$ where $\E_0$ is the true regression surface.

Often this function is assumed to be linear, in which case the goal is to find $\gamma \in \mathbb{R}^{p\times d}$, where $d < p$, such that $Y$ depends on $X$ only through $X^T \gamma$. That is,
\begin{align}
\E_0[Y \mid X] = \E[Y \mid X^T \gamma].
\label{eq:SDR}
\end{align}
$d$ is often referred to as the structural dimension. In a seminal paper,  \cite{ma2012semiparametric} recast the estimation problem in SDR as an estimation problem in a \emph{semiparametric} model.  This yielded a set of estimators of $\gamma$ that do not rely on strong assumptions on the observed data distribution.  This approach has since been extended to causal inference problems with high dimensional treatments \cite{nabi2017semi}.

Semiparametric estimators draw inferences from iid realizations
$Z_1, \ldots, Z_n$
drawn from a general class of probability densities $p(Z; \theta)$ parameterized by $\theta^T = (\beta^T, \eta^T)$, where $\beta \in \mathbb{R}^q$ denotes the (finite dimensional) set of target parameters, and $\eta$ denotes a possibly infinite dimensional set of nuisance parameters.  This type of model is termed semiparametric, since it has both a parametric and a non-parametric component.
\cite{tsiatis2007semiparametric,bickel1993efficient} describe a geometric approach to performing estimation in such settings, using the Hilbert space of scores of the model.  In such a framing, semiparametric estimators for $\beta$ with attractive properties correspond
to elements of the orthogonal complement of the nuisance tangent space of the model.
A detailed review of this approach can be found in the Supplement.

\section{The Interpretable Model}
\label{sec:inter_ML}
In this section we define our model, henceforth referred to as the IML (Interpretable Machine Learning) model.

Assume we are interested in learning the regression function $\E_0[Y | X]$ from realizations of the observed data distribution $p_0(Y, X)$, where $Y$ is a continuous valued outcome, $X$ is partitioned into a (small) set of interpretable features $X_{\text{int}}$, and a (large) set of uninterpretable features $X_{\text{uint}}$ of size $p$.  Our semiparametric model requires there exists a matrix $\gamma$ of size $p \times d$ such that
$Y = h(X_{\text{int}}; \psi) + r(X_{\text{uint}}^T \gamma) + \epsilon$ where $\mathbb{E}[\epsilon \mid X_{\text{int}}, X_{\text{uint}}] = 0$, and $\mathbb{E}[Y \mid X_{\text{int}}, X_{\text{uint}}] = \mathbb{E}[Y \mid X_{\text{int}}, X_{\text{uint}}^T\gamma]$.
No other restrictions are placed on the model.  We denote our model (the set of distributions with above restrictions) by ${\cal M}_{\text{int}}$.

In words, ${\cal M}_{\text{int}}$ assumes the regression surface we wish to learn is representable as a simple function $h()$ of interpretable features $X_{\text{int}}$ parameterized by $\psi$, and a complicated function $r()$ of a projection of $X_{\text{uint}}$ from its original dimension $p$ to a smaller dimension $d$ - referred henceforth as the structural dimension - chosen by the user, as given by the shape of $\gamma$.  Importantly, $r()$ is completely unrestricted.

We derive the space of influence functions yielding RAL estimators for $\beta \equiv (\psi, \gamma)$ by deriving $\Lambda^{\perp}$, the orthogonal complement of the nuisance tangent space of ${\cal M}_{\text{int}}$. \\

\begin{theorem}
	The set of all influence functions, i.e. the orthogonal complement of the nuisance tangent space, for $\beta$ in ${\cal M}_{\text{int}}$ is given as follows,
	{\small
	\begin{align*}
	\Lambda^{^{\perp}}_{\text{int}} &= 
	\Big\{  \big( A(X_{\text{int}}, X_{\text{uint}}) - \E[A(X_{\text{int}}, X_{\text{uint}}) \mid X^T_{\text{uint}}\gamma ] \big) \times \big( Y - h(X_{\text{int}};\psi) - r(X_{\text{uint}}^T\gamma) \big)   \Big\}
	\end{align*}
	}%
	where $A(X_{\text{int}}, X_{\text{uint}})$ is any $|\beta|$-dimensional function of $\{X_{\text{int}}, X_{\text{uint}}\}$. 
	\label{thm:if-c}
\end{theorem}
Given an arbitrary element $\phi_A(\beta) \in \Lambda^{\perp}_{\text{int}},$ we get a consistent RAL estimator by solving the estimating equation of the form $\E[ \phi_A(\beta) ] = 0$. The variance of such an estimator is given by the variance of $\phi_A(\beta)$. 

According to Theorem~\ref{thm:if-c}, the class of our estimators may be viewed as augmenting a particular parametric model $h(.;\psi)$ with a flexible, uninterpretable piece using a modification of semiparametric SDR.  We obtain a version of the double robustness result in \cite{ma2012semiparametric}, provided $h(.;\psi)$ itself is specified correctly.

\begin{lemma}
Given $\phi_A(\beta) \in \Lambda^{^{\perp}}_{\text{int}}$, an estimator for $\beta$ which solves $\E[ \phi_A(\beta) ] = 0,$ is consistent and asymptotically normal if $h(.;\psi)$ is correctly specified, and either of the models in $\{ r(X_{\text{uint}}^T \gamma), \ \mathbb{E}[A({X}) \mid X_{\text{uint}}^T \gamma]\}$ is correctly specified.
\label{lem:dr}
\end{lemma}

Lemma \ref{lem:dr} allows us to perform consistent inferences for $\beta$ even in settings where a large part of the model likelihood is arbitrarily misspecified, provided $h(.)$ and one of two models
are specified correctly.  In addition, double robustness implies the bias of the estimator has a product form.  This allows parametric ($\sqrt{n}$) convergence rates for $\beta$ to be obtained even if flexible machine learning models with slower than parametric convergence rates are used to fit nuisance models.  See \cite{chernozhukov2018double} for details.

Our semiparametric estimators, derived from Theorem \ref{thm:if-c}, may be extended to arbitrary link functions as well, and the influence functions are presented below. Denote the model ${\cal M}_{\text{g,\text{int}}}$ to be one where $Y = g(h(X_{\text{int}};\psi) + r(X_{\text{uint}}^T \gamma)) + \epsilon$, where $\mathbb{E}[\epsilon \mid X_{\text{int}}, X_{\text{uint}}] = 0$, and $g$ denotes a known differentiable link function.
\begin{theorem}
The set of all influence functions, i.e. the orthogonal complement of the nuisance tangent space, for $\beta$ in ${\cal M}_{\text{g,\text{int}}}$ is given as follows,
\begin{align*}
\Lambda^{\perp}_{\text{g,\text{int}}} \!=\! \{{A(X_{\text{int}},X_{\text{uint}}) - \mathbb{E}[A(X_{\text{int}},X_{\text{uint}}) | X_{\text{uint}}^T \gamma]} \} \times (g^\prime)^{-1} \times \{Y - g(h(X_{\text{int}};\psi) + r(X_{\text{uint}}))\}
\end{align*}
where $g^\prime$ is the derivative of the link function with respect to its single input.
\label{thm:if-d}
\end{theorem}
Choosing appropriate link functions allows us to consider dichotomous outcomes, often leading to submodels of ${\cal M}_{\text{g,\text{int}}}$.  Nevertheless, elements in $\Lambda^{\perp}_{\text{g,\text{int}}}$ remain consistent in such models.


Estimating equations of the form $\E[ \phi_A(\beta) ] = 0$ obtained from Theorem~\ref{thm:if-c} and \ref{thm:if-d} are not linear in $\beta$.
As a result, similarly to estimating equations corresponding to Z-estimators \cite{van2000asymptotic}, they must be solved numerically.
We utilize constrained non-linear optimization algorithms to learn our parameters, along with non-parametric kernel regressions to perform prediction using these learned parameters.
A detailed description of these algorithms are given in the Supplement.

In addition, our results so far assumed the structural dimension $d$ is fixed and known in advance.  In practice, this is an unrealistic assumption, and $d$ must be chosen according to some criterion.
We adapted a dimension selection procedure described in \cite{ma2012semiparametric, dongDimensionReductionNonelliptically2010}, with details left to the Supplement.



\section{Experiments}
\label{sec:exp}

We now demonstrate the performance of our estimator using various simulations, as well as a data application.

\subsection{Simulation Study}
In our simulation study, the performance is evaluated using two metrics, the first being the root mean squared error (RMSE) calculated on a held-out testing set. Next, we assess the model on parameter recovery, which is evaluated by the Frobenious norm of $\hat{\gamma}(\hat{\gamma}^T\hat{\gamma})^{-1}\hat{\gamma}^T - \gamma(\gamma^T\gamma)^{-1}\gamma^T$ and $\hat{\psi}(\hat{\psi}^T\hat{\psi})^{-1}\hat{\psi}^T - \psi(\psi^T\psi)^{-1}\psi^T$. This quantity ranges from 0 to 2, and the smaller the number, the better the estimate with 0 denoting a perfect match. 

To benchmark our model, we compare the performance of our model against a generalized additive model (GAM) \cite{hastie2017generalized} and a linear regression model. The GAM implementation used the \texttt{pygam} implementation with the \emph{n\_splines} parameter set to 25. 
We utilize parametric bootstrap and generate 50 replicates of the dataset, each of size $n = 2000$. The comparison is given in the tables below.

To simulate our data, we closely follow the simulations used in \cite{ma2012semiparametric}. We set $\psi = [0.577, -0.577]$, $\gamma_1 = [.4082, .4082, .4082, .4082,  .4082, .4082]$ and $\gamma_2 = [ .4082 , -.4082, .4082, -.4082, .4082, -.4082]$. We generate the covariates $X$ in two different ways, split across case 1 and case 2.

\textbf{Case I}: $X_{\text{int}}$ is generated from $\mathcal{N}(0, \Sigma)$, where $\Sigma$ is given by $0.5^{|a - b|}$ where a and b are the row and column indices respectively. $X_{\text{uint}}$ is generated from $\mathcal{N}(0, 3 {I})$. 

\textbf{Case II}: Similar to \cite{ma2012semiparametric}, $X_{\text{uint},1}, X_{\text{uint},2}$ are jointly generated from a multivariate normal distribution with mean 0 and covariance matrix given as $0.5^{|a - b|}$. $X_{\text{uint}, 3} = |X_{\text{uint},1} + X_{\text{uint},2}| + |X_{\text{uint}, 1}|\epsilon_1$ and $X_{\text{uint},4} = |X_{\text{uint},1} + X_{\text{uint},2}|^2 + |X_{\text{uint},2}|\epsilon_2$. $X_{\text{uint},5}$ comes from a Bernoulli distribution with $p = \frac{\exp(X_{\text{uint},2})}{1 + \exp(X_{\text{uint},2})}$. $X_{\text{uint},6}$ comes from $\phi(X_{\text{uint},2})$, where $\phi$ denotes the cumulative density function of the standard normal distribution. For the interpretable features, $X_{\text{int},1}, X_{\text{uint},2}$ are generated from a multivariate normal distribution with mean 0 and covariance matrix given as $0.5^{|a - b|}$. Then, $X_{\text{int},2}$ is transformed as $X_{\text{int},2}|X_{\text{uint},3}|$. 
Assume the following. \\
\begin{minipage}{.5\linewidth}
{\small
\begin{align*}
\epsilon &\sim \mathcal{N}(0, 1), \\
h(X_{\text{int}};\psi) &\equiv X_{\text{int}}^{T}\psi,  \\
r_1(X_{\text{uint}}^T\gamma) &= (X_{\text{uint}}^T\gamma_1)^2 + (X_{\text{uint}}^T\gamma_2)^2, \\ r_2(X_{\text{uint}}^T\gamma) &= \frac{(X_{\text{uint}}^T\gamma_1)}{(0.5 + (X_{\text{uint}}^T\gamma_2 + 1.5)^2)}. \\
\end{align*}
}%
\end{minipage}
\begin{minipage}{.5\linewidth}
{\small
\begin{align*}
\textrm{Model(I)}: Y &= h(X_{\text{int}};\psi) +  r_1(X_{\text{uint}}^T\gamma) + \epsilon\\
\textrm{Model(II)}: Y &= h(X_{\text{int}};\psi) +  r_1(X_{\text{uint}}^T\gamma) 
\\ \hspace{2.5cm} &+  |X_{\text{uint}}^T\gamma_1||X_{\text{int},1}|\epsilon\\
\textrm{Model(III)}: Y &= h(X_{\text{int}};\psi) +  r_2(X_{\text{uint}}^T\gamma) + \epsilon\\ 
\textrm{Model(IV)}:  Y &= h(X_{\text{int}};\psi) +  r_2(X_{\text{uint}}^T\gamma) + |X_{\text{uint}}^T\gamma_1|\epsilon.
\end{align*}
} %
\end{minipage}

The models we use in this simulation study belong to the class of multiple-index models \cite{multipleindex, ichimura1991semiparametric}, a flexible class of models that represent common models such as the partially linear model \cite{speckman1988kernel, wang2003dimension} and the single index model \cite{ichimura1991semiparametric}.  Parameters of models in this class are known to be difficult to estimate\cite{li2000efficient}.

{\tiny
\begin{center}
	\begin{table}[t]
		\caption{RMSE comparison between IML, linear regression, and GAM for Case I in simulated data.}
		\label{Case Results}
		\vskip 0.15in
		\begin{center}
			\begin{tiny}
				\begin{sc}
					\begin{tabular}{|c|ccc|ccc|}
						\multicolumn{1}{c}{}
						&
						\multicolumn{3}{c}{CASE I RMSE}
						&
						\multicolumn{3}{c}{CASE II RMSE}\\
						\toprule
						Model & IML & GLR & GAM &  IML & GLR & GAM\\
						\midrule
						 I    & \cellcolor{green} 1.722 & 6.064 & 5.029 & \cellcolor{green}9.903 & 16.149 & 10.41\\
						      &	 \cellcolor{green}(0.446) & (0.379) & 	(0.259) & 	\cellcolor{green}(4.766) &	 (9.533) & (9.28) \\ 	 
						 II   & \cellcolor{green}1.649 & 6.125 & 5.103 & 10.969 & 15.868 & \cellcolor{green}10.145\\
						      &\cellcolor{green}(0.386) & (0.402) & (0.311) & (7.56) &(5.225)& \cellcolor{green}(10.538) \\
						 III  &\cellcolor{green}1.125 & 1.436 & 1.392 &\cellcolor{green}1.719 & 2.171 & 1.729\\
						      & \cellcolor{green}(0.098) & (0.073) & (0.066) & \cellcolor{green}(0.225)& (0.13) & (0.216)\\
						 IV & \cellcolor{green}1.864 & 1.995 & 1.967 &\cellcolor{green} 3.642 & 3.686 & 7.052\\
						      &\cellcolor{green}(0.152)&(0.14)&(0.137)&\cellcolor{green}(0.456)&(0.399)& (7.688) \\
						\bottomrule
					\end{tabular}
				\end{sc}
			\end{tiny}
		\end{center}
		\vskip -0.1in
	\end{table}
\end{center}
}

\begin{figure*}
    \begin{subfigure}[b]{0.25\textwidth}
    \includegraphics[width = \linewidth]{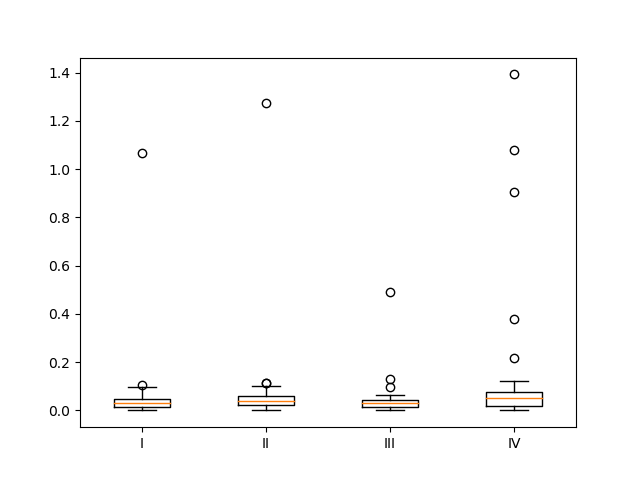}
    \caption{$\psi$ recovery in Case I}
    \end{subfigure}%
    \begin{subfigure}[b]{0.25\textwidth}
    \includegraphics[width= \linewidth]{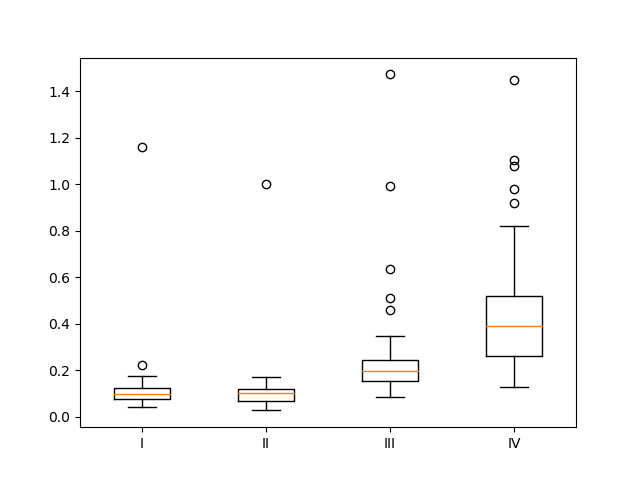}
    \caption{$\gamma$ recovery in Case I}
    \end{subfigure}%
    \begin{subfigure}[b]{0.25\textwidth}
    \includegraphics[width= \linewidth]{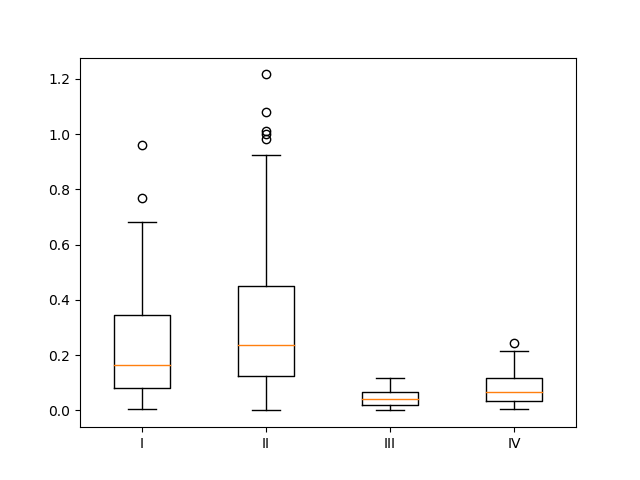}
    \caption{$\psi$ recovery in Case II}
    \end{subfigure}%
    \begin{subfigure}[b]{0.25\textwidth}
    \includegraphics[width= \linewidth]{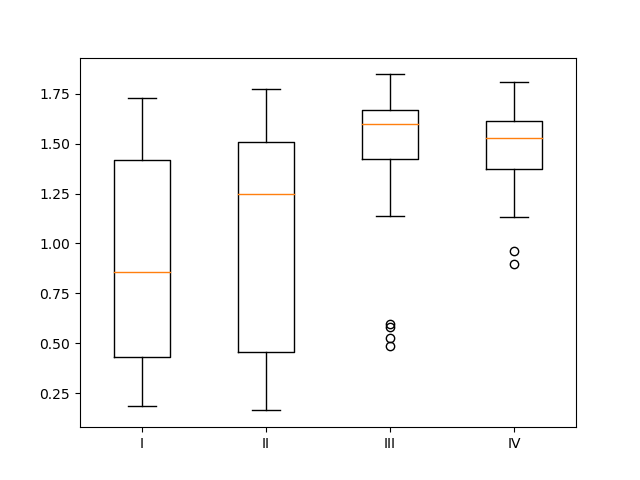}
    \caption{$\gamma$ recovery in Case II}
    \end{subfigure}
    \label{fig:eid}
    \caption{Boxplots demonstrating parameter recovery of our model: The X-axis in all of these graphs is the model number, while the Y-axis is the Frobenius norm of  $\hat{\gamma}(\hat{\gamma}^T\hat{\gamma})^{-1}\hat{\gamma}^T - \gamma(\gamma^T\gamma)^{-1}\gamma^T$ or $\hat{\psi}(\hat{\psi}^T\hat{\psi})^{-1}\hat{\psi}^T - \psi(\psi^T\psi)^{-1}\psi^T$, depending on the parameter it is for}
\end{figure*}
We used \texttt{scipy.optimize}, which supports non-convex optimization with non-linear constraints.  All simulations were run on the cluster at our institution, and took $50$ hours for $50$ bootstrap replicates at $n=2000$.

Tables \ref{Case Results} and \ref{ICU Data Application Results} represent the results.  The IML column lists the results from our interpretable model, the LR column lists the results from a Gaussian linear regression and the GAM columns lists the results from a generalized additive model. As we can see, our model outperforms both Gaussian linear regression as well as a GAM on the RMSE, except for in Table 2 model III, where the GAM is marginally better. Since the GAM is a more flexible model than the Gaussian linear regression, it performs better on the data. In addition to comparing performance on the mean squared error, we also measure our algorithm on parameter recovery and demonstrate the results in Figure 1.  The quality of the learned parameters is reasonable, given the sample size ($n=2000$), and the fact that the data generating process belongs to the class of multiple index models, known to be challenging\cite{li2000efficient}.

\subsection{Data Application}

In this section, we are interested in predicting ICU length of stay, as motivated by discussion in Section \ref{sec:tradeoff}.

\subsubsection{Partitioning Between Interpretable and Uninterpretable Features} \label{sec:int_uint}
An existing approach utilized for this problem is to compute a Parsonnet score \cite{lawrenceParsonnetScoreGood2000}, which calculates post-operative risk based on commonly available variables, and utilize it to predict LOS \cite{Lawrence429}. Variables utilized in this scoring system include gender, obesity, diabetes, hypertension, ejection fraction, age, number of previous operations, use of intra-aortic balloon pump, previous left ventricular aneurysm, dialysis, previous valve surgery, previous catastrophic state (cardiogenic shock, renal failure). The features used for this score were hand picked by clinicians from available EHR sources, and represent features that are of interest to clinicians in this domain. Hence, we utilized a subset of these features as the interpretable features in our model.

\cite{doeringDeterminantsIntensiveCare2001} found that factors associated with prolonged ICU stays in patients undergoing coronary artery bypass grafts (CABGs) also include recent myocardial infarction, smoking, diseased arteries, and preoperative left-ventricular diastolic pressure, as well as postoperative factors such as arrhythmias, respiratory complications, and renal insufficiency. This motivated us to pick uninterpretable features such as - various measures of blood transfusions at different phases of the clinical stay, reintubation status, postoperative creatinine level (a proxy for kidney function), perfusion time, ICU readmission, hematocrit level prior to surgery and platelet count.

Note that some of our uninterpretable features may have scientific interpretations and could in theory be assigned as interpretable features. However, we have restricted the set of interpretable features to those that are clinically interpretable according to the Parsonnet score, which includes features that clinicians are used to interpreting in the context of ICU stay prediction \cite{doeringDeterminantsIntensiveCare2001}. Our set of uninterpretable features consists of features which are associated with prolonged ICU stays, but are not part of the Parsonnet feature set. Hence, the interpretable feature set acts as a modified (yet interpretable) Parsonnet score, and the uninterpretable features act to improve the Parsonnet prediction performance.


\subsubsection{Data Preprocessing}
The electronic health records of 6189 cardiac surgery patients at a major research hospital were queried. The outcome variable is defined as the initial duration of the initial ICU stay, measured in hours from admittance to discharge from ICU. Patients with ICU stays in the top 2.5\% were excluded (as these patients are likely systematically different rom those normally entering the ICU), as were patients that did not enter the ICU at all. then, we discarded features and rows with missing data. This left us with a dataset of 5665 rows and 268 features. 9 interpretable features were selected based on their utilization in the Parsonnet score. An additional 10 uninterpretable features were selected based on discussion in Section \ref{sec:int_uint}. A detailed description of each feature is provided in the Supplement.
The Supplement
also demonstrates the performance on the prediction task using a different partition of the feature set.

\subsubsection{Benchmarking Performance}
We benchmarked our method against linear regression, random forests (RFs), and GAMs, and predictive performance was measured by root mean squared error (RMSE). To compare interpretability, we compared the coefficients from the linear regression model with the coefficients from our IML model. The random forest regressor from \texttt{sklearn} was run with various tuning parameters, with the number after RF indicating the n\_trees hyperparameter settings. The GAM from \texttt{pygam} was fitted with \emph{n\_splines} parameter set to 25. For our IML model, we utilized $\delta = 0.05$ and the structural dimension $d = 1$. Additional models of RFs and GAMs with different hyperparameter settings can be found in the Supplement,
as well as details with the training and validation accuracy.

The dataset was trained on a training dataset consisting of 3965 rows and 19 features, and a validation dataset of 851 rows was utilized. The performance of the various algorithms was compared using the predictions on a held-out testing set of size 849. The table above gives the train, validation and test RMSEs. As seen in Table \ref{Case Results}, we outperform all the different methods when it comes to RMSE, with RFs Forests coming in second.

	\begin{table}[t]
		\caption{RMSE comparison between IML, linear regression, GAM, and random forest using ICU data.}
		\label{ICU Data Application Results}
		\vskip 0.15in
		\begin{center}
			\begin{tiny}
				\begin{sc}
					\begin{tabular}{|c|c|c|c|}
						\toprule
						Model & Train & Val & Test \\
						\midrule
						IML & 49.734 & 57.793 & 53.748\\
						LR  & 51.147 & 57.808 & 55.93\\
						GAM25 & 48.773 & 60.925 & 56.46\\
						RF100 & 45.617 & 57.126 & 54.562\\
						\bottomrule
					\end{tabular}
					\quad
					\begin{tabular}{|c|c|c|}
						\toprule
						Feature & IML & LR \\
						\midrule
						Gender    &   -6.097 & -6.201\\
						Diabetes  & 5.95 & 5.867\\
						Hypertension  &1.794 & -1.884\\
						Pre-op IABP & 14.73 & 14.541\\
						Status:Emergent & 12.446 & 12.775\\
						 Dialysis & -22.618 &  -22.915\\
						 Carshock & 18.605 &18.126\\
						 Prvalve & 4.135& 4.3 \\
						 Prcab & 9.501 & 9.7383\\
						\bottomrule
					\end{tabular}
				\end{sc}
			\end{tiny}
		\end{center}
		\vskip -0.1in
	\end{table}

\subsubsection{Performance of Parametric and IML Models}
\begin{figure*}[t!]
    \centering
    \begin{subfigure}[t]{.45\textwidth}
        \includegraphics[width = \linewidth]{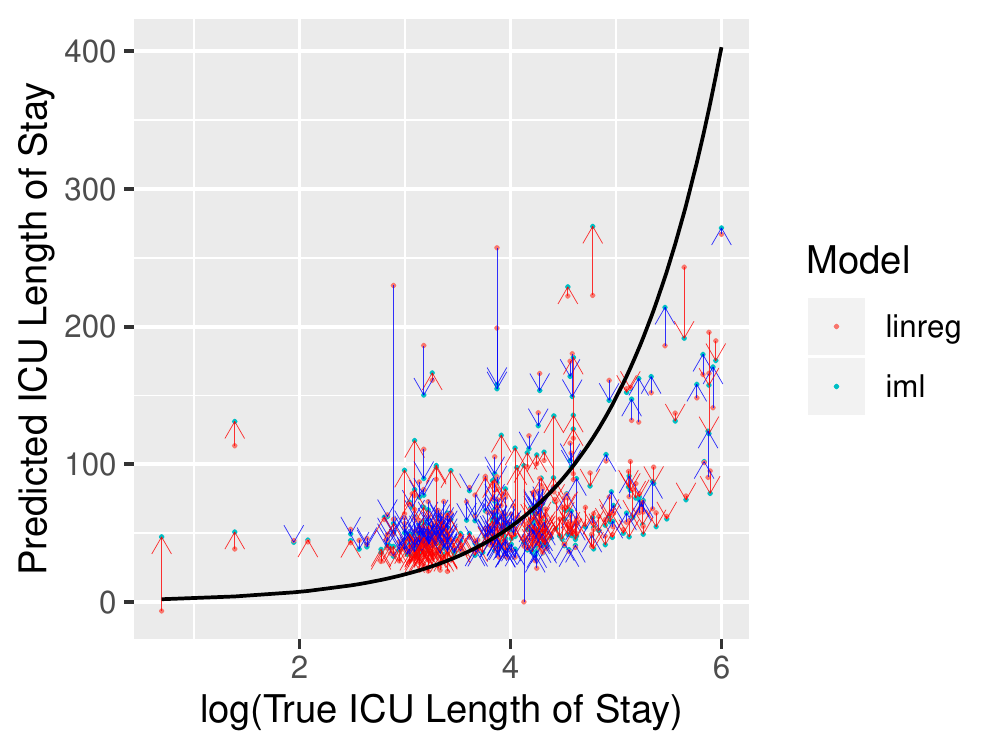}
        \caption{Comparison of Linear Regression (red) versus Interpretable ML (blue) Predictions}
        \label{fig:linreg_vs_iml}
    \end{subfigure}%
    ~
    \begin{subfigure}[t]{.45\textwidth}
        \includegraphics[width= \linewidth]{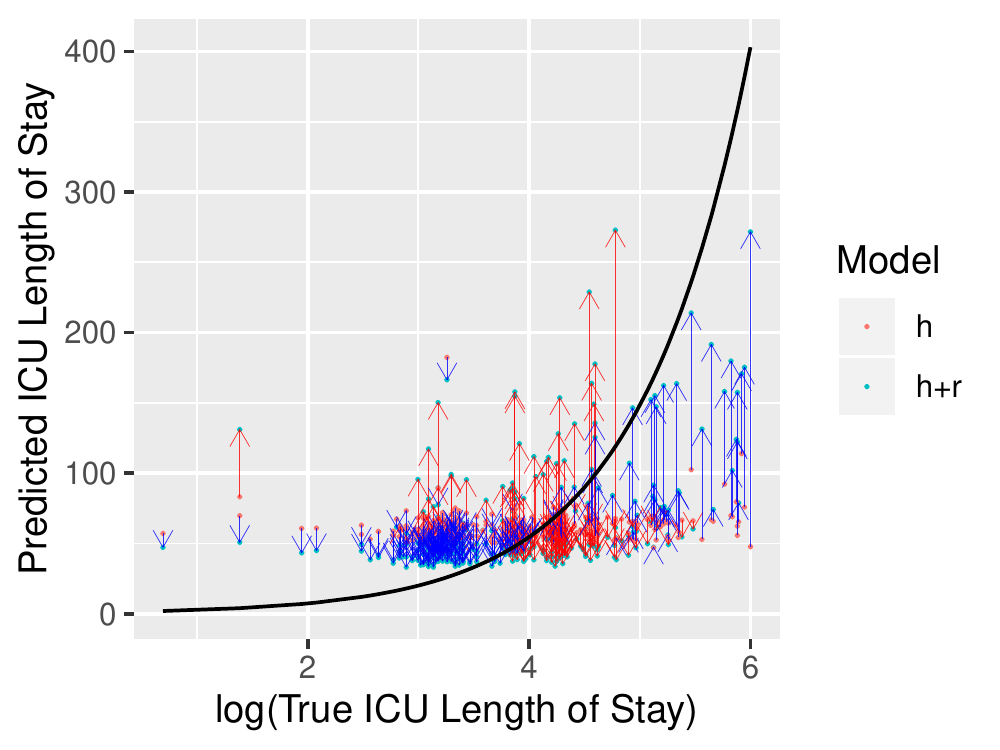}
        \caption{Comparison of the Interpretable $h()$ Function (red) versus $h() + r()$ Function (blue) Predictions}
        \label{fig:h_vs_hr}
    \end{subfigure}%
    \caption{The x-axis represents the natural logarithm of the true ICU stay duration, whilst the y-axis represents the predicted ICU stay duration. The black line are the set of points representing predictions exactly equal to the true value. In each plot, the comparison is made of the change of the blue model against the original red model. Blue arrows indicate that the blue model improved the prediction by bringing it closer to the black line, whilst red arrows indicate that the blue model worsened the prediction by moving it away. The length of the arrow indicates the degree of improvement/worsening. A random subsample of 400 patients was selected to improve visual clarity.}
\end{figure*}

IML outperforms all of the methods it is benchmarked against on the testing set, whilst keeping the coefficients of the interpretable features close to that of a linear regression. Figure \ref{fig:linreg_vs_iml} compares the performance of linear regression against IML. In most cases, we see that the IML prediction (blue) improves upon the linear regression prediction (red), as evidenced by the number of long blue arrows. The IML prediction only marginally worsens some of the linear regression predictions, evidenced by the number of small red arrows.

\begin{figure*}
    \begin{centering}
            \includegraphics[width = .6\linewidth]{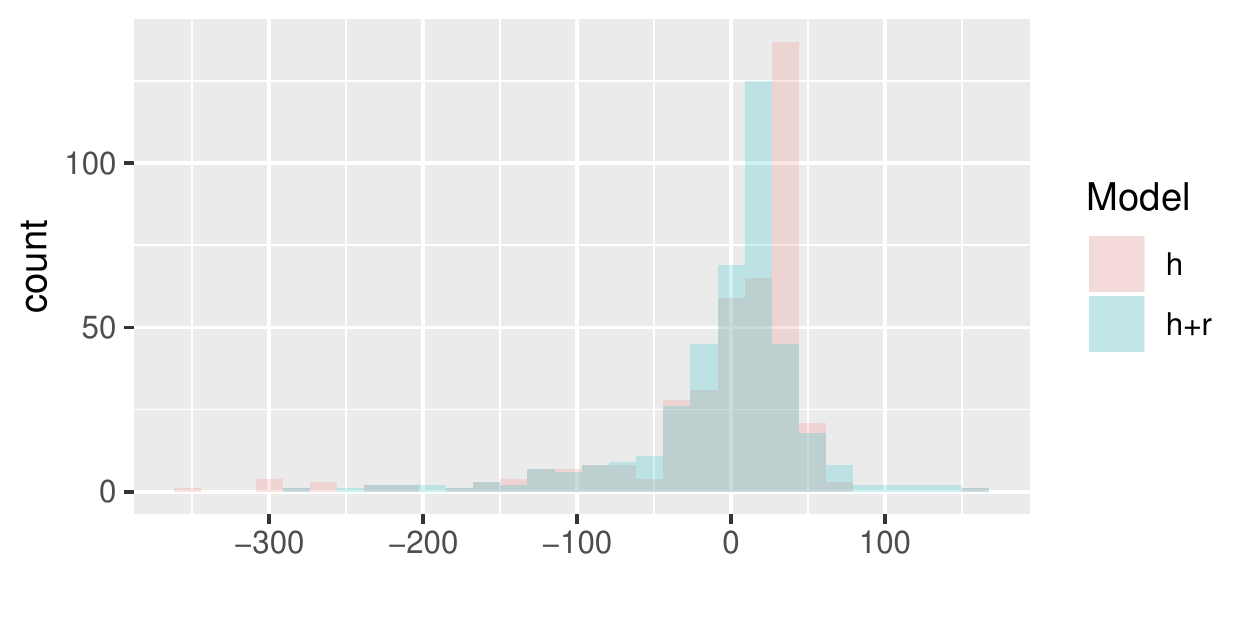}
            \caption{Histogram of residuals of the $h()$ function versus the $h() + r()$ functions.}
            \label{fig:h_vs_hr_hist}
    \end{centering}
\end{figure*}

While the IML model delivers greater predictive power while maintaining interpretability for some features, it is natural to consider the impact of the uninterpretable $r()$ function on the overall prediction. Figure \ref{fig:h_vs_hr} considers the impact of the $h()$ function alone (in red) compared to the $h() + r()$ function in blue. Primarily, the $r()$ function appears to improve upon the $h()$ function prediction for the lower and higher ICU durations, where a great number of blue arrows are evidenced.
Predictions for ICU durations close to the median were equal or slightly worse on average, with only a few instances with larger error.
Overall, the $r()$ function clearly improves the overall fit, as shown in Figure \ref{fig:h_vs_hr_hist}, and Table \ref{ICU Data Application Results}.

\section{Conclusion}
\label{sec:conc}
In this paper we presented a novel approach to interpretable machine learning based on semiparametric statistics and sufficient dimension reduction. Our approach began with an interpretable parametric model that used interpretable features as inputs, and added to it an unrestricted low dimensional function of a large set of uninterpretable features in a way that best mimicked the underlying regression surface.

This problem was posed as a target parameter learning problem in a semiparametric model.  The estimator for needed parameters was obtained by deriving the space of all influence functions for the target parameter.  The resulting estimators enjoyed a double robustness property, provided the parametric form of the interpretable part of the regression is known.
We have demonstrated, via simulation studies and a data application, the performance of our method, as well as its ability to retain much of the interpretable structure of a parametric regression model.

Deriving the efficient influence function for our problem, as well as adapting ideas from the sparsity literature to generalize this work to very high dimensional problems are promising areas of future work.

\section{Supplement}
\subsection{Inference in Semiparametric Models}
\label{sec:semi}

Let $Z_1, \ldots, Z_n$, be iid samples from a general class of probability densities $p(Z; \theta)$ parameterized by $\theta^T = (\beta^T, \eta^T)$, where $\beta \in \mathbb{R}^q$ denotes the (finite dimensional) set of target parameters, and $\eta$ denotes a possibly infinite dimensional set of nuisance parameters.  This type of model is termed semiparametric, since it has both a parametric and a non-parametric component.  The goal of statistical inference in semiparametric models is to find ``best" (lowest asymptotic variance) estimator of $\beta$ in the model, denoted by $\widehat{\beta}$.
We will consider \emph{regular asymptotically linear (RAL)} estimators, which are estimators of the form
\begin{eqnarray}
	\sqrt{n}(\widehat{\beta} - \beta) = \frac{1}{\sqrt{n}} \sum_{i = 1}^{n} \phi(Z_i) + o_p(1), \nonumber 
\end{eqnarray} 
where $\phi(Z_i)$ is the \emph{influence function (IF)} of the $i$th observation for the parameter vector $\beta$, and $o_p(1)$ denotes a term that approaches to zero in probability. The influence function for the parameter $\beta$ is a random function of the same dimension as $\beta$, i.e., $\phi(Z) \in \mathbb{R}^q$. Moreover, the IF is always mean zero, with certain regularity assumptions on the model ensuring that the IF has finite variance. 
RAL estimators are consistent and asymptotically normal, with the variance of the estimator given by the variance of its IF.
\begin{eqnarray}
	\sqrt{n}(\widehat{\beta} - \beta) \xrightarrow[]{\mathcal{D}} \mathcal{N}(0, \phi\phi^T). \nonumber 
\end{eqnarray}
Thus, there is a bijective correspondence between RAL estimators and IFs.  In fact, IFs provide a geometric view of the behavior of RAL estimators.

Consider a Hilbert space ${\cal H}$ of all mean-zero $q-$dimensional functions, equipped with an inner product. Define the inner product of two arbitrary elements of the Hilbert space, $h_1$ and $h_2$, as $\mathbb{E}[h_1^Th_2]$.
Define a \emph{parametric submodel} to be a subset of densities in the semiparametric model parameterized by $\theta^T_{\gamma} = (\beta^T, \gamma^T)$, where $\gamma^T \in {\mathbb R}^r$, such that the subset contains the density $p(Z; \theta_0)$ in the semiparametric model evaluated at the true parameter values $\theta_0$.
The \emph{nuisance tangent space} $\Lambda$ in the semiparametric model is defined to be the mean square closure of elements of the nuisance tangent spaces $\Lambda_{\gamma} = \{B^{q\times r} S_\gamma(Z; \theta) \}$ of every parametric submodel.

The space $\Lambda$ is important because it is known that all influence functions lie in the orthogonal complement of $\Lambda$, denoted by $\Lambda^{\perp}$,  with respect to ${\cal H}$ (${\cal H} = \Lambda \oplus \Lambda^\perp$, where $\oplus$ is the direct sum).  
For this reason, recovering $\Lambda^{\perp}$ is often the first step for constructing RAL estimators in semiparametric models.  Out of all IFs in
$\Lambda^{\perp}$ there exists a unique one which lies in the tangent space, and which yields the most efficient RAL estimator by recovering the
\emph{semiparametric efficiency bound}; see \cite{tsiatis2007semiparametric} for details.

A well-known property of semiparametric models is that all elements of $\Lambda^{\perp}$ are mean $0$ under the true distribution. Consequently, given an arbitrary element $U(\beta) \in \Lambda^\perp,$ we get an estimating equation of the form $\E[U(\beta)] = 0$ which upon solving, yields a consistent and asymptotically linear estimator for $\beta.$

\subsection{Sufficient Dimension Reduction}

The sparsity literature \cite{hastie15statistical} places strong assumptions on the regression surface allowing procedures that pick a small subset of the overall feature set to optimize prediction, even with very few samples relative to the size of the feature set.  Sparsity approaches to regression problems are generally parametric.

A common approach to reducing a high-dimensional feature set to a low dimensional one is dimension reduction approaches such as principal component analysis (PCA), that aims to select a set of features that captures major axes of variation given by the feature covariance matrix \cite{pearson01on}.  The difficulty with methods based on the PCA for regression problems is that they either ignore the outcome entirely, or take no special precautions to preserve the feature outcome relationship.  This means that regression models built on top of the output of PCA type procedures may perform significantly worse compared to those built on the original set of high dimensional features.
A popular class of methods that may be viewed as a dimension-reduction strategy that take the feature outcome relationships into account, such as linear discriminant analysis (LDA) and generalizations often make strong parametric assumptions.

A powerful strategy that avoids the problems described above, and allows the reduction of the dimension of the features while preserving a given relationship between features and the outcome \emph{exactly} has been developed in the \emph{sufficient dimension reduction (SDR)} literature in statistics. 

Let $X$ be a $p$-dimensional covariate vector and $Y$ be a univariate response variable. The goal of SDR is to find a known function $r(.; \gamma)$, with a much smaller range than the domain, parameterized by $\gamma$ such that $\E_0[Y \mid X] = \E[Y \mid r(X; \gamma)],$ where $\E_0$ is the true regression surface.

Often this function is assumed to be linear, in which case the goal is to find $\gamma \in \mathbb{R}^{p\times d}$, where $d < p$, such that $Y$ depends on $X$ only through $X^T \gamma$. That is,
\begin{align}
	\E_0[Y \mid X] = \E[Y \mid X^T \gamma].
	\label{eq:SDR}
\end{align}
There are various effective methods developed over the course of decades for estimating the set of matrices $\gamma$.

Examples include sliced inverse regression \cite{li1991sliced}, sliced average variance estimation \cite{cook1991sliced}, and directional regression \cite{li2007directional}. 
However,  all of these approaches rely on strong parametric assumptions that are unlikely to hold in practical applications.   Commonly used in the literature are assumptions such as the linearity assumption -$\E[X \mid  X^T \gamma]$ is a linear function of $X$, or the constant variance assumption - $\text{cov}(X \mid X^T \gamma)$ is a constant rather than a function of $X$.

In a seminal paper,  \cite{ma2012semiparametric} recast the estimation problem in SDR as an estimation problem in a \emph{semi-parameteric} model.  This yielded a set of estimators of $\gamma$ that do not rely on strong assumptions on the observed data distribution.  This approach has since been extended to causal inference problems with high dimensional treatments \cite{nabi2017semi}.
\cite{ma2012semiparametric} derived the orthocomplement of the nuisance tangent space for the model in (\ref{eq:SDR}):
{\small
	\begin{align}
		\Lambda^{\perp} = \big\{ \big(Y - \E[Y \mid X^T \beta] \big) \times \big( \alpha(X) - \E[\alpha(X) \mid X^T\beta]\big) \big\},
		\label{eq:MA_orthocomp}
	\end{align}
}%
where $\alpha(X)$ is an arbitrary function of $X$ of an appropriate dimension.  Thus, a consistent family of estimators for $\beta$ (for different $\alpha(X)$) for semiparametric sufficient dimension reduction is:

{\small
	\begin{align}
		\E\big[ \big(Y - \E[Y \mid X^T \beta] \big) \times \big( \alpha(X) - \E[\alpha(X) \mid X^T\beta]\big) \big] = 0.
		\label{eq:MA_SDR}
	\end{align}
}%
\cite{ma2012semiparametric} illustrated how other parametric SDR methods are special cases of the above family of semiparametric estimators. For example, ordinary least squares (OLS) \cite{li1989regression} can be viewed as a special case of the above estimator. To obtain this, set $\alpha(X) = X$ and take advantage of the double robustness of the above estimator to set $\mathbb{E}[Y \mid X^T\beta] = 0$. This combined with the linearity assumption of the OLS estimator yields the estimating equations for OLS.

The estimator in (\ref{eq:MA_SDR}) is \textit{doubly robust} with respect to models for $\E[Y \mid X^T \beta]$ and $\E[\alpha(X) \mid X^T \beta]$, meaning that the estimator remains consistent if any \emph{one} of these two models is correctly specified (even if the other model is arbitrarily misspecified).

If the influence function is linear in $\beta$, then $\beta$ can be estimated in closed form after all nuisance models are estimated.

However, estimating equations such as (\ref{eq:MA_SDR}) entail solving for $\beta$ using iterative methods, such as variants of the Newton-Raphson algorithm.

\subsection{Proofs For Section 4}
\subsubsection{Proof of Theorem 1}
For our model where
\begin{align*}
	&Y = h(X_{int};\psi) + l(X_{uint}^T\gamma) + \epsilon\\
	&\mathbb{E}[\epsilon \mid X_{int}, X_{uint}] = 0
\end{align*}
and we observe i.i.d data in the form ${Z} = (Y, X_{int}, X_{uint})$, the goal is to find estimators for $\langle \psi, \gamma \rangle$. The semiparametric model is $\{p({z}; \beta, \eta)\: \beta, \eta\}$ where $\beta$ is a finite dimensional parameter and $\eta = \langle \psi, \gamma \rangle$ is an infinite dimensional parameter. We let $p_0({z})$ denote the true data distribution, and $\beta_0, \eta_0$ denote the true value of the parameters.

The likelihood for our model is given as:
\begin{align*}
	p(Y, X_{int}, X_{uint}) &= p(Y \mid X_{int}, X_{uint})p(X_{uint}, X_{int})\\
	&= p(\epsilon \mid X_{int}, X_{uint})p(X_{uint}, X_{int})\\
	&= p(\epsilon \mid X_{int}, X_{uint}; \eta_2)p(X_{uint}, X_{int}; \eta_1)
\end{align*}
where $\epsilon = Y - h(X_{int};\psi) - l(X_{uint}^T\gamma)$. Here $\eta_1, \eta_2, l$ are infinite dimensional nuisance parameters.

The nuisance tangent space $\Lambda$ is the space spanned the nuisance score vectors. 
\begin{align*}
	\Lambda &= \{B \times S_\eta \forall B\}\\
	S_\eta &= \frac{\partial \log p({z}; \beta, \eta)}{\partial \eta}\\
	&= \{\frac{\partial \log p(X_{uint}, X_{int}; \eta_1)}{\partial \eta_1},\frac{\partial \log p(\epsilon \mid X_{int}, X_{uint}; \eta_2)}{\partial \eta_2} ,\frac{\partial \log p(\epsilon \mid X_{int}, X_{uint}; \eta_2)}{\partial l}\}
\end{align*}

$S_{\eta_1}$ and $S_{\eta_2}$ are score vectors, they must be mean zero, and we have the added restriction $\mathbb{E}[\epsilon \mid X_{int}, X_{uint}] = 0$. Following an argument similar to \cite{tsiatis2007semiparametric}, we obtain:
\begin{align*}
	\Lambda_{\eta_1} &= \{A(X_{int}, X_{uint}) : \mathbb{E}[A(X_{int}, X_{uint})] = 0\} \\
	\Lambda_{\eta_2} &= \{\alpha(\epsilon, X_{int}, X_{uint}) : \mathbb{E}[\alpha(\epsilon, X_{int}, X_{uint}) \mid X_{uint}, X_{int}] = 0, \\
	&\hspace{0.5in}\mathbb{E}[\alpha(\epsilon, X_{int}, X_{uint})\epsilon^T \mid X_{uint}, X_{int}] = 0\} \\
	\Lambda_{l} &= \{\frac{\partial p(\epsilon \mid X_{int}, X_{uint}; \eta_2)/\partial \epsilon}{p(\epsilon \mid X_{int}, X_{uint}; \eta_2)}m(X_{uint}^T\gamma) : \forall m\}
\end{align*}

A detailed description of the derivation of $\Lambda_{\eta_1}$ can be found in Theorem 4.6 in \cite{tsiatis2007semiparametric}. Similarly, the derivation of derivation of $\Lambda_{\eta_2}$ can be found in Theorem 4.7 in \cite{tsiatis2007semiparametric}. The derivation of $\Lambda_l$ follows a similar argument to \cite{nabi2017semi}, with a brief overview given below:
\begin{align*}
	S_l &= \frac{\partial \log p(y - h - l \mid X_{int}, X_{uint}; \eta_2)}{\partial l}\\
	&= \frac{\partial p(\epsilon = y - h - l \mid X_{int}, X_{uint}; \eta_2)/ \partial \epsilon}{p(\epsilon = y - h - l \mid X_{int}, X_{uint}; \eta_2)}\frac{\partial \epsilon}{\partial l}\\
	&= \frac{\partial p(\epsilon = y - h - l \mid X_{int}, X_{uint}; \eta_2)/ \partial \epsilon}{p(\epsilon = y - h - l \mid X_{int}, X_{uint}; \eta_2)}\frac{\partial l}{\partial l}\\
	\Lambda_l &= \frac{\partial p(\epsilon \mid X_{int}, X_{uint}; \eta_2)/\partial \epsilon}{p(\epsilon \mid X_{int}, X_{uint}; \eta_2)}m(X_{uint}^T\gamma) : \forall m\}
\end{align*}

Since $l$ is allowed to belong to the unrestricted space of functions, it follows that its derivative will belong to the unrestricted space of functions as well. 
Since $\eta_1, \eta_2, l$ are variationally independent:
\begin{align*}
	\Lambda &= \Lambda_{\eta_1} \oplus \Lambda_{\eta_2} \oplus \Lambda_{l}\\
	\Lambda^\perp &= \Lambda_1^\perp \cap \Lambda_2^\perp \cap \Lambda_l^\perp
\end{align*}
As $\Lambda^\perp$ is the intersection of 3 spaces, $\Lambda^\perp \subset (\Lambda^\perp_{\eta_1} \cap \Lambda^\perp_{\eta_2}) =  (\Lambda_{\eta_1} \oplus \Lambda_{\eta_2})^\perp$, and $\Lambda^\perp \subset \Lambda^\perp_l$. Based on Theorem 4.8 of \cite{ma2012semiparametric}:
\begin{align*}
	(\Lambda_1 \oplus \Lambda_2)^\perp &= \{A(X_{int}, X_{uint})\epsilon : \forall A(X_{int}, X_{uint})\}
\end{align*}

So any element of $\Lambda^\perp$ must have the form above with additional restrictions being placed on it by the intersection with  $\Lambda_l^\perp$. Taking any element of $\Lambda^\perp$ of the form $\{A(X_{int}, X_{uint})\epsilon\}$, this must be orthogonal to $\Lambda_l$. This yields:
\begin{align*}
	\mathbb{E}[A(X_{int}, X_{uint})\epsilon \frac{\partial p(\epsilon \mid X_{int}, X_{uint})/\partial \epsilon}{p(\epsilon \mid X_{int}, X_{uint})}m(X_{uint}^T\gamma)] = 0
\end{align*}

Which implies that $\mathbb{E}[A(X_{int}, X_{uint})m(X_{uint}^T\gamma)] = 0, \forall m$. This only holds if $\mathbb{E}[A(X_{int}, X_{uint}) \mid X_{uint}^\gamma] = 0$. Hence, any element of $\Lambda_\perp$ can be written as:
\begin{align*}
	\Lambda^\perp = \{A(X_{int}, X_{uint}) - \mathbb{E}[A(X_{int}, X_{uint}) \mid X_{uint}^T\gamma]\}\{Y - h(X_{int}; \psi) - l(X_{uint}^T\gamma)\}
\end{align*}

\subsubsection{Proof of Theorem 2}
The derivation of the influence function in the presence of an arbitrary known link function follows a similar outline, except the $\Lambda_l$ space is different.
\begin{align*}
	\Lambda_l = \{\frac{\partial p(\epsilon \mid X_{int}, X_{uint})/ \partial \epsilon}{p(\epsilon \mid X_{int}, X_{uint})}g^\prime(h(X_{int};\psi) + l(X_{uint}^T\gamma))m(X_{uint}^T\gamma): \forall m\}
\end{align*}

Similar to before, using the orthogonality condition between $\Lambda^\perp$ and $\Lambda_l$:
\begin{align*}
	\mathbb{E}[A(X_{int}, X_{uint})\epsilon \frac{\partial p(\epsilon \mid X_{int}, X_{uint})/\partial \epsilon}{p(\epsilon \mid X_{int}, X_{uint})}g^\prime(h(X_{int};\psi) + l(X_{uint}^T\gamma))m(X_{uint}^T\gamma)] = 0
\end{align*}
This gives us the additional restriction on $A$,
\begin{align*} 
	\mathbb{E}[A(X_{int}, X_{uint})g^\prime(h(X_{int};\psi) + l(X_{uint}^T\gamma))m(X_{uint}^T\gamma)] = 0 
\end{align*}
Taking this into account, $\Lambda^\perp$ is given below.
\begin{align*}
	\Lambda^\perp = \{\frac{A(X_{int},X_{uint}) - \mathbb{E}[A(X_{int},X_{uint}) \mid X_{uint}^T\gamma]}{g^\prime} \}\{Y - g(h(X_{int};\psi) - l(X_{uint}))\}
\end{align*}

\subsubsection{Proof of Lemma 1}
Let $\mathbb{E}^*[A(X_{int}, X_{uint}) \mid X_{uint}^T\gamma]$ denoted the incorrectly specified model, yielding the following estimating equation:
\begin{align*}
	\mathbb{E}[\{A(X_{int}, X_{uint}) - \mathbb{E}^*[A(X_{int}, X_{uint}) \mid X_{uint}^T\gamma]\}\\
	\times \{Y - h(X_{int}; \psi) - r(X_{uint}^T\gamma)\}]
\end{align*}
Utilizing the law of iterated expectation, we have
{\small
	\begin{align*}
		\mathbb{E}_{{X}}[& \mathbb{E}_{Y \mid {X}}[\{A(X_{int}, X_{uint}) - \mathbb{E}^*[A(X_{int}, X_{uint}) \mid X_{uint}^T\gamma]\}\\
		&\times \{Y - h(X_{int}; \psi) - r(X_{uint}^T\gamma)\} \mid X_{int}, X_{uint}]]\\
		= \mathbb{E}_{{X}}[&\{A(X_{int}, X_{uint}) - \mathbb{E}^*[A(X_{int}, X_{uint}) \mid X_{uint}^T\gamma]\}\{\mathbb{E}[Y \mid {X}] - h(X_{int}; \psi) - r(X_{uint}^T\gamma)\}]\\
	\end{align*}
}
But $\mathbb{E}[Y \mid {X}] = h(X_{int}; \psi) + r(X_{uint}^T\gamma)$, so the above expectation will evaluate to $0$, despite the incorrectly specified model.

Instead, now assume that the $r(X_{uint}^T\gamma)$ model is incorrectly specified, denoted by $r^*$. The estimating equation is then given as:
\begin{align*}
	\mathbb{E}[\{A(X_{int}, X_{uint}) - \mathbb{E}[A(X_{int}, X_{uint}) \mid X_{uint}^T\gamma]\}\\
	\times \{Y - h(X_{int}; \psi) - r^*(X_{uint}^T\gamma)\}]
\end{align*}
Using iterated expectation, we obtain:
{\small
	\begin{align*}
		\mathbb{E}_{{X}}[\mathbb{E}_{Y \mid X}[\{A(X_{int}, X_{uint}) - \mathbb{E}[A(X_{int}, X_{uint}) \mid X_{uint}^T\gamma]\}\{Y - h(X_{int}; \psi) - r^*(X_{uint}^T\gamma)\}\mid {X}]]\\
		= \mathbb{E}_{{X}}[\{A(X_{int}, X_{uint}) - \mathbb{E}[A(X_{int}, X_{uint}) \mid X_{uint}^T\gamma]\}\{\mathbb{E}_{Y \mid X}[Y] - h(X_{int}; \psi) - r^*(X_{uint}^T\gamma)\}]\\
		= \mathbb{E}_{{X}}[\{A(X_{int}, X_{uint}) - \mathbb{E}[A(X_{int}, X_{uint}) \mid X_{uint}^T\gamma]\}\{r(X_{uint}^T\gamma)- r^*(X_{uint}^T\gamma)\}]
	\end{align*}
}
Now taking iterated expectation again:
{\small
	\begin{align*}
		\mathbb{E}_{X_{uint}^T\gamma}[& \mathbb{E}_{X_{int} \mid X_{uint}^T\gamma}[\{A(X_{int}, X_{uint}) - \mathbb{E}[A(X_{int}, X_{uint}) \mid X_{uint}^T\gamma]\}\\
		\times & \{r(X_{uint}^T\gamma)- r^*(X_{uint}^T\gamma)\} \mid X_{uint}^T\gamma]]\\
		= \mathbb{E}_{X_{uint}^T\gamma}[\{&\mathbb{E}_{X_{int} \mid X_{uint}^T\gamma}[A(X_{int}, X_{uint})\mid X_{uint}^T\gamma] - \mathbb{E}[A(X_{int}, X_{uint}) \mid X_{uint}^T\gamma]\}\\
		\times &\{r(X_{uint}^T\gamma)- r^*(X_{uint}^T\gamma)\}]\\
	\end{align*}
}
This will also evaluate to $0$, as long as the model $\mathbb{E}[A(X_{int}, X_{uint}) \mid X_{uint}^T\gamma]$ is correct, even if $r^*$ is not specified correctly.

We have shown that as long as the parametric form of $h$ is known, the estimating equation for $\beta$ evaluates to $0$ with respect to the observed data distribution, as long as either $\mathbb{E}[A(X_{int}, X_{uint}) \mid X_{uint}^T\gamma]$ or $r$ is specified correctly.  This establishes double robustness.

\subsection{Fitting Procedure}

In this section, we describe in detail our procedure for estimating $\beta$ by solving the empirical version of the estimating equation, $\widehat{E}[\phi_A(\beta)] = 0$, where $\phi_A(\beta) \in \Lambda^{^{\perp}}_{\text{int}}.$

Without loss of generality, assume $\psi \in \mathbb{R}^{p\times 1}$ and $\gamma \in \mathbb{R}^{q\times d},$ where $d$ denotes the structural dimension of the lower dimensional representation of $X_{\text{uint}}$ (we delay the discussion on the choice of $d$ to further sections.) For a given choice of $A(X) \equiv A(X_{\text{int}}, X_{\text{uint}})$ and $h(X_{\text{int}}; \psi)$,

\begin{enumerate}
	
	\item Pick starting values for $\beta^{(1)} = \{\psi^{(1)}, \gamma^{(1)}\}$. To initialize $\psi,$ we fit a linear regression of $Y$ on $X$ and use the coefficients learned for $X_{\text{int}}$. To initialize $\gamma,$  we use an SDR procedure called principal Hessian directions described in \cite{li1992principal}. 
	
	\item At $j$th iteration, given a fixed $\beta^{(j)} = \{\psi^{(j)}, \gamma^{(j)}\}$, fit the following models. 
	
	{\scriptsize
		\begin{align*}
			&\mathbb{E}[Y \mid X_{k, \text{\text{uint}}}^T\gamma^{(j)}] = 
			\frac{\sum_{i = 1}^n Y_k \times K_h(X_{i, \text{\text{uint}}}^T\gamma^{(j)} - X_{\text{\text{uint}}}^T\gamma^{(j)})}{ \sum_{i = 1}^n K_h(X_{i, \text{\text{uint}}}^T\gamma^{(j)} - X_{\text{\text{uint}}}^T\gamma^{(j)})}
			\\
			&\mathbb{E}[A(X_{\text{\text{uint}}}, X_{\text{\text{int}}} ) \mid X_{\text{k, uint}}^T \gamma^{(j)}]  \\
			&\hspace{1cm} = \frac{\sum_{i = 1}^n A(X_{k, \text{\text{uint}}}, X_{k, \text{\text{int}}} ) \times K_h(X_{i, \text{\text{uint}}}^T\gamma^{(j)} - X_{\text{\text{uint}}}^T\gamma^{(j)})}{ \sum_{i =1}^{n} K_h(X_{i,\text{\text{uint}}}^T\gamma^{(j)} - X_{\text{\text{uint}}}^T\gamma^{(j)})}
			\\
			&\mathbb{E}[h(X_{\text{\text{int}}}; \psi^{(j)}) \mid X_{k, \text{\text{uint}}}^T\gamma]  \\
			&\hspace{1cm} = \frac{\sum_{i = 1}^n h(X_{k, \text{\text{int}}}; \psi^{(j)}) \times K_h(X_{i, \text{\text{uint}}}^T\gamma^{(j)} - X_{\text{\text{uint}}}^T\gamma^{(j)})}{ \sum_{i = 1}^n K_h(X_{i, \text{\text{uint}}}^T\gamma^{(j)} - X_{\text{\text{uint}}}^T\gamma^{(j)})},
			\\
			&r(X^T_{k, \text{\text{uint}}}\gamma^{(j)}) = \E[Y \mid X^T_{k, \text{\text{uint}}}\gamma^{(j)}]  - \mathbb{E}[h(X_{k, \text{\text{int}}}; \psi^{(j)}) \mid X_{k, \text{\text{uint}}}^T\gamma^{(j)}],
		\end{align*}
	}%
	
	where $K_h(.)$ denotes a kernel function with bandwidth $h$. We use a Gaussian kernel in our experiments. We estimate the bandwidth $h$ using the Silverman's Rule of Thumb \cite{lauter1988silverman}.

	\item Form the empirical evaluation of $\E[\phi_{A}(\beta^{(j)})]$ as, 
	{\small
		\begin{align*}
			&e(X; \psi^{(j)}, \gamma^{(j)}) \\
			&\hspace{0.8cm} \equiv \ \frac{1}{n} \sum_{i=1}^n \big\{A(X_i) - \mathbb{E}[A(X) \mid X_{i,\text{\text{uint}}}^T\gamma^{(j)}]\big\} \\
			&\hspace{1cm} \times \big\{Y_i - h(X_{i,\text{\text{int}}};\psi^{(j)}) -r(X^T_{i, \text{\text{uint}}}\gamma^{(j)})  \big\},
		\end{align*}
	}
	
	\item Using a non-convex optimizer along with the supporting non-linear constraints, minimize the Frobenius norm of $e(X; \psi^{(j)}, \gamma^{(j)}),$ subject to the following constraints. (Given a matrix $\beta$, Frobenius norm is defined as $\beta(\beta^T\beta)^{-1}\beta^T$)
	\begin{itemize}
		\item[(a)] $	\gamma^T\gamma - \mathbb{I} = 0, \quad \text{ s.t. } \quad -1 \leq \gamma \leq 1, $ \\
		
		\item[(b)] $| \psi  - \psi_{init} |_\textrm{max} \leq \delta $, where $\delta$ is a non-negative tuning parameter chosen by the analyst.
	\end{itemize}
\end{enumerate}

This restriction on $\gamma$ put forth in 4(a) is to ensure that among possible choices for $\gamma$ we prefer those that do well on the optimization objective while not leading to drastically different values of $\psi$ from those in the initial parametric model. The restriction in 4(b) allows control between a model which is more interpretable (where $\delta$ is close to 0) or which has better prediction performance (where $\delta$ is set away from 0).

A commonly used choice for $A(X_{\text{int}}, X_{\text{uint}})$ is a simple function of the appropriate dimension, e.g. $X_{\text{int}}, X_{\text{uint}}, X_{\text{uint}}^2 \dots$. This is what we use for our simulations and data application. There exists an \emph{optimal} choice of $A$ (in the sense of minimizing the variance of the estimated parameters $\beta$) that can be obtained by projecting $\Lambda^\perp$ on to the tangent space. We leave the choice of optimal $A$ to future work.

In practice, it is recommended to standardize the features $X_{\text{\text{int}}}, X_{\text{\text{uint}}}$ to have mean zero and unit variance. 

\subsubsection{Using the Learned Parameters for Prediction}

Having discussed the method to estimate the parameters, we now outline the procedure to use these estimated parameters for prediction on new data. Let $\widehat{\beta} = \{\widehat{\psi}, \widehat{\gamma}\}$ be the estimated parameters, and $X_j = \{X_{j, \text{\text{int}}} \cup X_{j, \text{\text{uint}}} \}$ be the new datapoint. In order to get a prediction of $Y$ for $X_j$, we use the following. 
{\small
	\begin{align*}
		\E[Y \mid X_j; \{ \widehat{\psi}, \widehat{\gamma} \} ] = h(X_{j, \text{\text{int}}}; \widehat{\psi}) + r(X_{j, \text{\text{uint}}}; \widehat{\gamma}), 
	\end{align*}
}%
where the form of $h(X_{j, \text{\text{int}}}; \widehat{\psi})$ is known and 
{\small
	\begin{align*}
		r(X_{j, \text{\text{uint}}}^T \widehat{\gamma}) =  \E[Y \mid X^T_{k, \text{\text{uint}}}\widehat{\gamma}]  - \mathbb{E}[h(X_{j, \text{\text{int}}}; \widehat{\psi}) \mid X_{j, \text{\text{uint}}}^T\widehat{\gamma}],
	\end{align*}
}
The contribution of the uninterpretable features to the prediction is not as straightforward, since $r$ is a complex function learned from the data using non-parametric methods (kernels in our case). We learn the $r(.)$ function from the data by fitting the following non-parametric kernel regressions with $i$ ranging over the training data.
{\small
	\begin{align*}
		&\mathbb{E}[Y \mid X_{j,\text{uint}}^T\hat{\gamma}] = 
		\frac{\sum_{i = 1}^n Y_i \times K_h(X_{i,\text{uint}}^T\hat{\gamma} - X_{j,\text{uint}}^T\hat{\gamma})}{ \sum_{i = 1}^n K_h(X_{i,\text{uint}}^T\hat{\gamma} - X_{j,\text{uint}}^T\hat{\gamma})}
		\\
		&\mathbb{E}[h(X_{\text{int}};\hat{\psi}) \mid X_{j,\text{uint}}^T\hat{\gamma}] 
		\\
		&\hspace{2cm} = \frac{\sum_{i = 1}^n h(X_{i,\text{int}}; \hat{\psi}) \times K_h(X_{i,\text{uint}}^T\hat{\gamma} - X_{j,\text{uint}}^T\hat{\gamma})}{\sum_{i = 1}^n K_h(X_{i,\text{uint}}^T\hat{\gamma} - X_{j,\text{uint}}^T\hat{\gamma})}
	\end{align*}
}
Evaluating the $r$ function gives the contribution of the uninterpretable features to the prediction. The prediction for the new point is the sum of the $h$ and $r$ function.

\subsubsection{Choosing  the Dimension}
\label{subsec:dim}
In the description of the fitting procedure above, the structural dimension of our estimator $d$ (the rank of $\gamma$) was considered fixed and known in advance.
In practice, $d$ is a meta-parameter that must be determined from data.  For this purpose, we modified the dimension selection procedure outlined in \citet{ma2012semiparametric, dongDimensionReductionNonelliptically2010}.

Let $\lambda_1, \ldots, \lambda_k$ be the non-zero eigenvalues of 
\[\var(u)^{-1/2} \cov(u, v^T) \var(v)^{-1} \cov(u, v^T) \var(u)^{-1/2}\]
for generic random vectors $u, v$. Then, define 
\[r^2 (u, v) = \frac{1}{k} \sum_{i=1}^k \lambda_i\]
For any structural dimension $k$, define $\hat{\gamma}_k$ to be the estimate based on $X$, and $\hat{\gamma}_{k, b}$ to be the estimate based on the $b$-th bootstrap sample of $X_{\text{uint}}$ for $b = 1, \ldots, B$. This returns a score 
\[\bar{r}^2_k = \frac{1}{B} \sum_{b=1}^B r^2(\hat{\gamma}_k^T X_{\text{uint}}, \hat{\gamma}_{k,b}^T X_{\text{uint}}),\]
which is to be maximized over candidate dimensions $k$, in order to determine the optimal dimension $d$.





\subsection{Data Application Details}

\subsubsection{Feature Descriptions}
Given the prediction task of ICU length of stay prediction, the first step is to motivate a split of features between those deemed interpretable, versus uninterpretable. To motivate the interpretable features, we rely on those involved in the Parsonnet score \cite{parsonnetMethodUniformStratification1989}, and label others as uninterpretable. Thus, the proposed IML model can be interpreted as an improved Parsonnet score.

The Parsonnet scoring system relies the following 17 features: gender, obesity, diabetes, hypertension, ejection fraction, age, reoperation status, preoperative IABP, left ventricular aneurysm, emergent status, dialysis, catastrophic state (proxied by cardiogenic shock), other rare circumstances (assortment of rare conditions such as paraplegia, severe asthma, etc.), past valve surgery, mitral valve disease, aortic valve disease, and previous CABGs.

Of these, our institutional database contained information on all above variables except left ventricular aneurysm, reoperation status, other rare circumstances, and mitral/aortic valve diseases/pressures. Hence, we are able to evaluate approximately the Parsonnet score on our database.

Due to reasons of computational resources, we selected from the Parsonnet score 9 interpretable features: gender, diabetes, hypertension, pre-operative intra-aortic balloon pump usage, emergent status, dialysis, cardiogenic shock, previous valve operation, and previous coronary artery bypass. The 10 uninterpretable features, selected from other non-Parsonnet EHR features, were: postoperative platelet units used, postoperative red blood cell units used, reintubation status, postoperative creatinine level, perfusion time, hematocrit, intraoperative blood products used, preoperative white blood cell count, previous cardiac intervention, and whether an ICU readmission occurred during the visit.

\subsubsection{Benchmark Model Details}
We benchmark our model against a Random Forest and a GAM. For the Random Forest, we utilized the \textit{sklearn.ensemble} package and set the max depth to 5 and number of trees to 100. For the GAM, we used the \textit{pygam} package and we utilize 25 splines along with the grid search option.

\subsection{Alternative Analysis in Data Application}


Instead of selecting interpretable features based directly on the Parsonnet score, we can instead select interpretable features from those known to be associated with the target \cite{almashrafiSystematicReviewFactors2016}, and uninterpretable features from those suspected of being associated with the target. 6 interpretable features were selected based on their appearance in previous scoring systems and relevance to the domain, and 10 additional uninterpretable features were selected based on feature information as determined by a random forest model. 

The interpretable features include measures of hypertension, previous cardiac interventions, weight, gender, age, and white blood cell count (a proxy for immune system performance). Uninterpretable features included various measures of blood transfusions at different phases of the clinical stay, reintubation status, postoperative creatinine level (a proxy for kidney function), perfusion time, use of an intra-aortic balloon pump, hematocrit level prior to surgery, use of antibiotics, and platelet count.

We used root mean squared error (RMSE) was used to benchmark our estimator against linear regression, random forests, and GAMs. The random forest regressor had tuning parameter \emph{n\_trees} set to 25. The GAM implementation used the \texttt{pygam} implementation with the \emph{n\_splines} parameter set to 25, along with the use of the grid search option. The dataset was trained on a training dataset consisting of 3965 rows and 16 features, and a validation dataset of 851 rows was utilized. The performance of the various algorithms was compared using the predictions on a held-out testing set of size 849. The table above gives the train, validation and test RMSEs. 
\begin{table}[t]
	\caption{RMSE comparison between IML, linear regression, GAM, and random forest using ICU data.}
	\label{ICU Data Application Results}
	\vskip 0.15in
	\begin{center}
		\begin{small}
			\begin{sc}
				\begin{tabular}{cccc}
					\toprule
					& Train & Validation & Test \\
					\midrule
					IML    & 51.033 & 57.068 & 54.314\\
					LR   & 51.442 & 57.532 & 56.05\\
					GAM  & 49.607 & 57.376 & 55.030\\
					RF & 21.58 & 57.532 & 54.664\\
					\bottomrule
				\end{tabular}
			\end{sc}
		\end{small}
	\end{center}
	\vskip -0.1in
\end{table}

\begin{table}[t]
	\caption{Comparing IML coefficients vs. linear regression coefficients.}
	\label{Coefficient Comparison}
	\vskip 0.15in
	\begin{center}
		\begin{small}
			\begin{sc}
				\begin{tabular}{ccc}
					\toprule
					Feature & IML & LR \\
					\midrule
					hypertension    & -2.84 & -2.724\\
					prev cardiac interven   & 6.695 & 6.802\\
					age  &0.013 & 0.012\\
					weight &0.061 & 0.058\\
					wbc count & 1.395 & 1.407\\
					gender & -6.096 & -6.181\\
					\bottomrule
				\end{tabular}
			\end{sc}
		\end{small}
	\end{center}
	\vskip -0.1in
\end{table}
\pagebreak



\bibliography{references}

\begin{thebibliography}{10}

\bibitem{almashrafiSystematicReviewFactors2016}
Ahmed Almashrafi, Mustafa Elmontsri, and Paul Aylin.
\newblock Systematic review of factors influencing length of stay in {{ICU}}
  after adult cardiac surgery.
\newblock {\em BMC Health Services Research}, 16, July 2016.

\bibitem{alvarez2018towards}
David Alvarez-Melis and Tommi~S Jaakkola.
\newblock Towards robust interpretability with self-explaining neural networks.
\newblock In {\em Proceedings of the 32nd International Conference on Neural
  Information Processing Systems}, pages 7786--7795. Curran Associates Inc.,
  2018.

\bibitem{bickel1993efficient}
Peter~J Bickel, Chris~AJ Klaassen, Peter~J Bickel, Ya’acov Ritov, J~Klaassen,
  Jon~A Wellner, and YA'Acov Ritov.
\newblock {\em Efficient and adaptive estimation for semiparametric models},
  volume~4.
\newblock Johns Hopkins University Press Baltimore, 1993.

\bibitem{opacityml}
Jenna Burrell.
\newblock How the machine ‘thinks’: Understanding opacity in machine
  learning algorithms.
\newblock {\em Big Data \& Society}, 3(1):2053951715622512, 2016.

\bibitem{chernozhukov2018double}
Victor Chernozhukov, Denis Chetverikov, Mert Demirer, Esther Duflo, Christian
  Hansen, Whitney Newey, and James Robins.
\newblock Double/debiased machine learning for treatment and structural
  parameters, 2018.

\bibitem{choi2016retain}
Edward Choi, Mohammad~Taha Bahadori, Jimeng Sun, Joshua Kulas, Andy Schuetz,
  and Walter Stewart.
\newblock Retain: An interpretable predictive model for healthcare using
  reverse time attention mechanism.
\newblock In {\em Advances in Neural Information Processing Systems}, pages
  3504--3512, 2016.

\bibitem{cook1991sliced}
R~Dennis Cook and Sanford Weisberg.
\newblock Sliced inverse regression for dimension reduction: Comment.
\newblock {\em Journal of the American Statistical Association},
  86(414):328--332, 1991.

\bibitem{doeringDeterminantsIntensiveCare2001}
Lynn~V Doering, Fardad Esmailian, Flerida {Imperial-Perez}, and Sheri Monsein.
\newblock Determinants of intensive care unit length of stay after coronary
  artery bypass graft surgery.
\newblock 30(1):9, 2001.

\bibitem{dongDimensionReductionNonelliptically2010}
Y.~Dong and B.~Li.
\newblock Dimension reduction for non-elliptically distributed predictors:
  Second-order methods.
\newblock {\em Biometrika}, 97(2):279--294, June 2010.

\bibitem{doshivelez2017rigorous}
Finale Doshi-Velez and Been Kim.
\newblock Towards a rigorous science of interpretable machine learning, 2017.

\bibitem{du2018techniques}
Mengnan Du, Ninghao Liu, and Xia Hu.
\newblock Techniques for interpretable machine learning.
\newblock {\em arXiv preprint arXiv:1808.00033}, 2018.

\bibitem{fong2017interpretable}
Ruth~C Fong and Andrea Vedaldi.
\newblock Interpretable explanations of black boxes by meaningful perturbation.
\newblock In {\em Proceedings of the IEEE International Conference on Computer
  Vision}, pages 3429--3437, 2017.

\bibitem{hastie15statistical}
Trevor Hastie, Robert Tibshirani, and Martin Wainwright.
\newblock {\em Statistical Learning With Sparsity: The Lasso and
  Generalizations}.
\newblock CRC Press, 2015.

\bibitem{hastie2017generalized}
Trevor~J Hastie.
\newblock Generalized additive models.
\newblock In {\em Statistical models in S}, pages 249--307. Routledge, 2017.

\bibitem{ichimura1991semiparametric}
Hidehiko Ichimura.
\newblock Semiparametric least squares (sls) and weighted sls estimation of
  single-index models.
\newblock 1991.

\bibitem{kramer2010predictive}
Andrew~A Kramer and Jack~E Zimmerman.
\newblock A predictive model for the early identification of patients at risk
  for a prolonged intensive care unit length of stay.
\newblock {\em BMC medical informatics and decision making}, 10(1):27, 2010.

\bibitem{lakkaraju2017interpretable}
Himabindu Lakkaraju, Ece Kamar, Rich Caruana, and Jure Leskovec.
\newblock Interpretable \& explorable approximations of black box models, 2017.

\bibitem{lauter1988silverman}
H~L{\"a}uter.
\newblock Silverman, bw: Density estimation for statistics and data analysis.
  chapman \& hall, london--new york 1986, 175 pp.,{\pounds} 12.—.
\newblock {\em Biometrical Journal}, 30(7):876--877, 1988.

\bibitem{lawrenceParsonnetScoreGood2000}
D~R Lawrence.
\newblock Parsonnet score is a good predictor of the duration of intensive care
  unit stay following cardiac surgery.
\newblock {\em Heart}, 83(4):429--432, April 2000.

\bibitem{Lawrence429}
D~R Lawrence, O~Valencia, E~E~J Smith, A~Murday, and T~Treasure.
\newblock Parsonnet score is a good predictor of the duration of intensive care
  unit stay following cardiac surgery.
\newblock {\em Heart}, 83(4):429--432, 2000.

\bibitem{lee2019intensive}
Hyun~Woo Lee, Yeonkyung Park, Eun~Jin Jang, and Yeon~Joo Lee.
\newblock Intensive care unit length of stay is reduced by protocolized family
  support intervention: a systematic review and meta-analysis.
\newblock {\em Intensive care medicine}, pages 1--10, 2019.

\bibitem{li2007directional}
Bing Li and Shaoli Wang.
\newblock On directional regression for dimension reduction.
\newblock {\em Journal of the American Statistical Association},
  102(479):997--1008, 2007.

\bibitem{li1991sliced}
Ker-Chau Li.
\newblock Sliced inverse regression for dimension reduction.
\newblock {\em Journal of the American Statistical Association},
  86(414):316--327, 1991.

\bibitem{li1992principal}
Ker-Chau Li.
\newblock On principal hessian directions for data visualization and dimension
  reduction: Another application of stein's lemma.
\newblock {\em Journal of the American Statistical Association},
  87(420):1025--1039, 1992.

\bibitem{li1989regression}
Ker-Chau Li and Naihua Duan.
\newblock Regression analysis under link violation.
\newblock {\em The Annals of Statistics}, pages 1009--1052, 1989.

\bibitem{li2000efficient}
Qi~Li.
\newblock Efficient estimation of additive partially linear models.
\newblock {\em International Economic Review}, 41(4):1073--1092, 2000.

\bibitem{lipton2016mythos}
Zachary~C. Lipton.
\newblock The mythos of model interpretability, 2016.

\bibitem{ma2012semiparametric}
Yanyuan Ma and Liping Zhu.
\newblock A semiparametric approach to dimension reduction.
\newblock {\em Journal of the American Statistical Association},
  107(497):168--179, 2012.

\bibitem{Murdoch_2019}
W.~James Murdoch, Chandan Singh, Karl Kumbier, Reza Abbasi-Asl, and Bin Yu.
\newblock Definitions, methods, and applications in interpretable machine
  learning.
\newblock {\em Proceedings of the National Academy of Sciences},
  116(44):22071–22080, Oct 2019.

\bibitem{nabi2017semi}
Razieh Nabi and Ilya Shpitser.
\newblock Semi-parametric causal sufficient dimension reduction of high
  dimensional treatments.
\newblock {\em arXiv preprint arXiv:1710.06727}, 2017.

\bibitem{parsonnetMethodUniformStratification1989}
V.~Parsonnet, D.~Dean, and A.~D. Bernstein.
\newblock A method of uniform stratification of risk for evaluating the results
  of surgery in acquired adult heart disease.
\newblock {\em Circulation}, 79(6 Pt 2):I3--12, June 1989.

\bibitem{jamiainterpretable}
Seyedeh~Neelufar Payrovnaziri, Zhaoyi Chen, Pablo Rengifo-Moreno, Tim Miller,
  Jiang Bian, Jonathan~H Chen, Xiuwen Liu, and Zhe He.
\newblock {Explainable artificial intelligence models using real-world
  electronic health record data: a systematic scoping review}.
\newblock {\em Journal of the American Medical Informatics Association}, 05
  2020.
\newblock ocaa053.

\bibitem{pearson01on}
K.~Pearson.
\newblock On lines and planes of closest fit to systems of points in space.
\newblock {\em Philosophical Magazine}, 2(11):559--572, 1901.

\bibitem{pronovost2006intervention}
Peter Pronovost, Dale Needham, Sean Berenholtz, David Sinopoli, Haitao Chu,
  Sara Cosgrove, Bryan Sexton, Robert Hyzy, Robert Welsh, Gary Roth, et~al.
\newblock An intervention to decrease catheter-related bloodstream infections
  in the icu.
\newblock {\em New England Journal of Medicine}, 355(26):2725--2732, 2006.

\bibitem{ribeiro2016i}
Marco~Tulio Ribeiro, Sameer Singh, and Carlos Guestrin.
\newblock "why should i trust you?": Explaining the predictions of any
  classifier, 2016.

\bibitem{simonyan2013deep}
Karen Simonyan, Andrea Vedaldi, and Andrew Zisserman.
\newblock Deep inside convolutional networks: Visualising image classification
  models and saliency maps, 2013.

\bibitem{speckman1988kernel}
Paul Speckman.
\newblock Kernel smoothing in partial linear models.
\newblock {\em Journal of the Royal Statistical Society: Series B
  (Methodological)}, 50(3):413--436, 1988.

\bibitem{tsiatis2007semiparametric}
Anastasios Tsiatis.
\newblock {\em Semiparametric theory and missing data}.
\newblock Springer Science \& Business Media, 2007.

\bibitem{van2000asymptotic}
Aad~W Van~der Vaart.
\newblock {\em Asymptotic statistics}, volume~3.
\newblock Cambridge university press, 2000.

\bibitem{wang2003dimension}
Qihua Wang.
\newblock Dimension reduction in partly linear error-in-response models with
  validation data.
\newblock {\em Journal of Multivariate Analysis}, 85(2):234--252, 2003.

\bibitem{pmlr-v97-wang19a}
Tong Wang.
\newblock Gaining free or low-cost interpretability with interpretable partial
  substitute.
\newblock In Kamalika Chaudhuri and Ruslan Salakhutdinov, editors, {\em
  Proceedings of the 36th International Conference on Machine Learning},
  volume~97 of {\em Proceedings of Machine Learning Research}, pages
  6505--6514, Long Beach, California, USA, 09--15 Jun 2019. PMLR.

\bibitem{wang2016bayesian}
Tong Wang, Cynthia Rudin, Finale Velez-Doshi, Yimin Liu, Erica Klampfl, and
  Perry MacNeille.
\newblock Bayesian rule sets for interpretable classification.
\newblock In {\em 2016 IEEE 16th International Conference on Data Mining
  (ICDM)}, pages 1269--1274. IEEE, 2016.

\bibitem{wold1987principal}
Svante Wold, Kim Esbensen, and Paul Geladi.
\newblock Principal component analysis.
\newblock {\em Chemometrics and intelligent laboratory systems}, 2(1-3):37--52,
  1987.

\bibitem{multipleindex}
Yingcun Xia.
\newblock A multiple-index model and dimension reduction.
\newblock {\em Journal of the American Statistical Association},
  103(484):1631--1640, 2008.

\end{thebibliography}
\bibliographystyle{plain}
\end{document}